\documentclass[letterpaper, 10 pt, conference]{ieeeconf}  

\usepackage[OT1]{fontenc}
\newcommand{\switchlanguage}[2]{%
  \ifx\paperlanguage\empty%
    #1%
  \else%
    #2%
  \fi%
}

\usepackage{amsmath,amssymb}
\usepackage{url,hyperref}
\usepackage{booktabs}
\usepackage{graphicx}
\usepackage{siunitx}
\usepackage{cite}
\usepackage{cases}
\usepackage{subcaption}
\usepackage{xcolor}

\newcommand{\tabref}[1]{{Table \ref{#1}}}
\newcommand{\figref}[1]{{Fig. \ref{#1}}}
\newcommand{\equref}[1]{{Eq. \ref{#1}}}

\newcommand{\bs}{\boldsymbol}

\IEEEoverridecommandlockouts

\def\paperlanguage{} 

\title{\LARGE \bf
{
    \switchlanguage%
    {%
    Construction of Musculoskeletal Simulation for Shoulder Complex with Ligaments and Its Validation via Model Predictive Control
    }%
    {%
    靭帯を含む肩複合体の筋骨格シミュレーションの構築とモデル予測制御による妥当性検証
    }%
  }
}

\author{Yuta Sahara$^{1}$, Akihiro Miki$^{1}$, Yoshimoto Ribayashi$^{1}$, Shunnosuke Yoshimura$^{1}$, \\Kento Kawaharazuka$^{1}$, Kei Okada$^{1}$, and Masayuki Inaba$^{1}$
\thanks{$^{1}$ The authors are with the Department of Mechano-Informatics, Graduate School of Information Science and Technology, The University of Tokyo, 7-3-1 Hongo, Bunkyo-ku, Tokyo, 113-8656, Japan.
  {\texttt\small [sahara, miki, ribayashi, yoshimura, kawaharazuka, k-okada, inaba]@jsk.t.u-tokyo.ac.jp}
}
}

\begin{document}

\maketitle
\thispagestyle{empty}
\pagestyle{empty}

\begin{abstract}
  \switchlanguage%
  {%
    The complex ways in which humans utilize their bodies in sports and martial arts are remarkable, and human motion analysis is one of the most effective tools for robot body design and control.
    On the other hand, motion analysis is not easy, and it is difficult to measure complex body motions in detail due to the influence of numerous muscles and soft tissues, mainly ligaments.
    In response, various musculoskeletal simulators have been developed and applied to motion analysis and robotics.
    However, none of them reproduce the ligaments but only the muscles, nor do they focus on the shoulder complex, including the clavicle and scapula, which is one of the most complex parts of the body.
    Therefore, in this study, a detailed simulation model of the shoulder complex including ligaments is constructed.
    The model will mimic not only the skeletal structure and muscle arrangement but also the ligament arrangement and maximum muscle strength.
    Through model predictive control based on the constructed simulation, we confirmed that the ligaments contribute to joint stabilization in the first movement and that the proper distribution of maximum muscle force contributes to the equalization of the load on each muscle, demonstrating the effectiveness of this simulation.
  }%
  {%
    スポーツや武術における人間の複雑な身体活用方法には目を見張るものがあり、
    ロボットの身体設計や制御を考えるうえで人間の運動解析は有効な手段の一つである。
    一方で運動解析は容易ではなく、多数の筋と靭帯を主とした軟部組織の影響により、
    その複雑な身体運動を詳細に計測することが難しい。
    これに対して、様々な筋骨格シミュレータが開発され、
    運動解析及びロボットへの応用が試みられてきた。
    しかし、それらのいずれも筋肉の再現のみで靭帯は再現しておらず、
    また、最も複雑な部位の一つである鎖骨や肩甲骨を含む肩複合体にも着目していない。
    そこで本研究では、靭帯を含む肩複合体の詳細なシミュレーションモデルを構築する。
    骨格や筋配置だけでなく、靭帯の配置や最大筋力まで模倣する。
    構築したシミュレーションに基づくモデル予測制御により、拳上動作において、
    靭帯が関節の安定化に寄与すること、
    最大筋力の適切な分配が各筋肉への負荷の均一化に寄与することを確認し、
    本シミュレーションの有効性を示した。
  }%
\end{abstract}

\section{Introduction}\label{sec:introduction}
\switchlanguage%
{%
  Humans make effective use of their bodies to achieve the complex movements seen in sports and martial arts.
  However, detailed analysis of how they move is difficult.
  This is because it is difficult to measure the contribution to movement of each of the more than 600 muscles in the human body.
  It is also difficult to measure the effects of soft tissues, mainly ligaments, which are greatly involved in joint movement.
  For these reasons, various musculoskeletal simulators have been developed and applied to motion analysis and robotics.
  However, existing musculoskeletal simulators have been constructed as simple models with a small number of muscles or without considering soft tissues such as ligaments.
  For example, in the model of Holzbaur et al.\cite{holzbaur2005model} the shoulder joint is configured as a joint that satisfies the regression equation obtained by de Groot et al.\cite{ de2001three} by experimentally studying the shoulder bone interlocking, with 16 muscles belonging to the shoulder complex in one shoulder.
  In myoSuite's myoArm, the shoulder is configured as a multiple uniaxial rotational joint with 63 muscles in the entire single arm, of which 15 muscles belong to the shoulder complex in one shoulder\cite{MyoSuite2022}.
  There are also Kengoro\cite{asano2016kengoro},
  musculoskeletal humanoids, which are wire-driven robots that mimic human muscles.
  Kengoro does not have ligaments in the shoulder joints and must have tension on the surrounding muscles or it will dislocate.
  Furthermore, Musashi\cite{kawaharazuka2019musashi} is a humanoid that includes the advantages of wire drive in musculoskeletal humanoids such as Kengoro, but it does not have scapula on its shoulders and does not adopt the structure of the human shoulder.

  In this study, a new musculoskeletal simulation model with soft tissue ligaments and the same number of muscles as a human was developed in order to apply the human shoulder structure, including its soft tissues and numerous muscles, to a musculoskeletal robot.
  In particular, the shoulder complex, which has the widest range of motion due to the function of muscles and ligaments, was the main target of the development.
  Ligaments were introduced to the joints in the same arrangement as in the human body, and 20 muscles were placed in one shoulder, for a total of 40 muscles involved in driving the shoulder complex.
  The maximum strength of each muscle was estimated from the volume and length of the muscle.
  In this paper, the effect of the ligaments was verified by performing raising movement experiments with and without the ligaments, and the uniformity of the load on each muscle by distributing the maximum muscle force was verified by performing raising movement experiments when the maximum muscle force of each muscle was estimated and when it was averaged.
}%
{%
  人間は身体を有効に活用して、スポーツや武術に見られる複雑な動作を実現している。
  しかし、その動かし方に関する詳細な解析は困難である。
  なぜならば、人体に存在する600を超える筋肉一つ一つによる運動への寄与を測定することは難しいからである。
  また、関節運動に大きく関わる、靭帯を主とした軟部組織の影響も計測が困難だからである。
  そのため、これまで様々な筋骨格シミュレータが開発され、
  運動の解析およびロボット等への応用が試みられてきた。
  しかし、既存の筋骨格シミュレータは、筋数が少なく、
  あるいは靭帯のような軟部組織も考慮されておらず、
  簡単なモデルとして構成されてきた。
  例えば、Holzbaurらのモデルでは、de Grootらが肩の骨の連動性について実験的に調べることによって得られた回帰方程式を満たす関節として肩の関節を構成しており、肩複合体に属する筋肉は片方の肩で16である\cite{holzbaur2005model,de2001three}。
  MyoSuiteのmyoArmでは肩を複数の1軸回転関節として構成しておりその筋数は片腕全体で63であるが、
  そのうち肩複合体に属する筋肉は片方の肩で15である\cite{MyoSuite2022}。
  また、人間の筋肉を模倣したワイヤー駆動ロボットである筋骨格ヒューマノイドとして、Kengoro\cite{asano2016kengoro}
  がいる。
  Kengoroは肩の関節に靭帯を有しておらず、周辺の筋肉に張力をかけることで脱臼を防いでいる。
  さらに、Kengoroなどの筋骨格ヒューマノイドにおけるワイヤー駆動の良さを取り入れたヒューマノイドとしてMusashi\cite{kawaharazuka2019musashi}が挙げられるが、肩に肩甲骨が無く、人間の肩の構造を取り入れているわけではない。

  そこで、本研究では、人間の肩の構造をその軟部組織や多数の筋肉まで含めて筋骨格ロボットへ応用させるため、
  軟部組織である靭帯を有して、 筋数も人間と同等である、新たな筋骨格シミューションモデルを開発した。
  特に、筋や靭帯の働きによって最も可動域の広い、肩複合体を主な対象として開発を行った。
  関節に対し、人体と同等の配置で靭帯を導入し、肩複合体の駆動に関わる筋肉を片方の肩で20本、併せて40本配置した。
  それぞれの筋肉の最大筋力は、筋肉の体積と長さから推定した。
  本論文では、靭帯を有する場合とない場合での挙上動作実験を行うことで靭帯の効果を検証し、
  各筋肉の最大筋力を推定したときと平均化したときとで挙上動作実験を行うことで最大筋力の分配による各筋肉への負荷の均一化を検証した。
}%

\begin{figure}[t]
  \centering
  \includegraphics[width=1\linewidth]{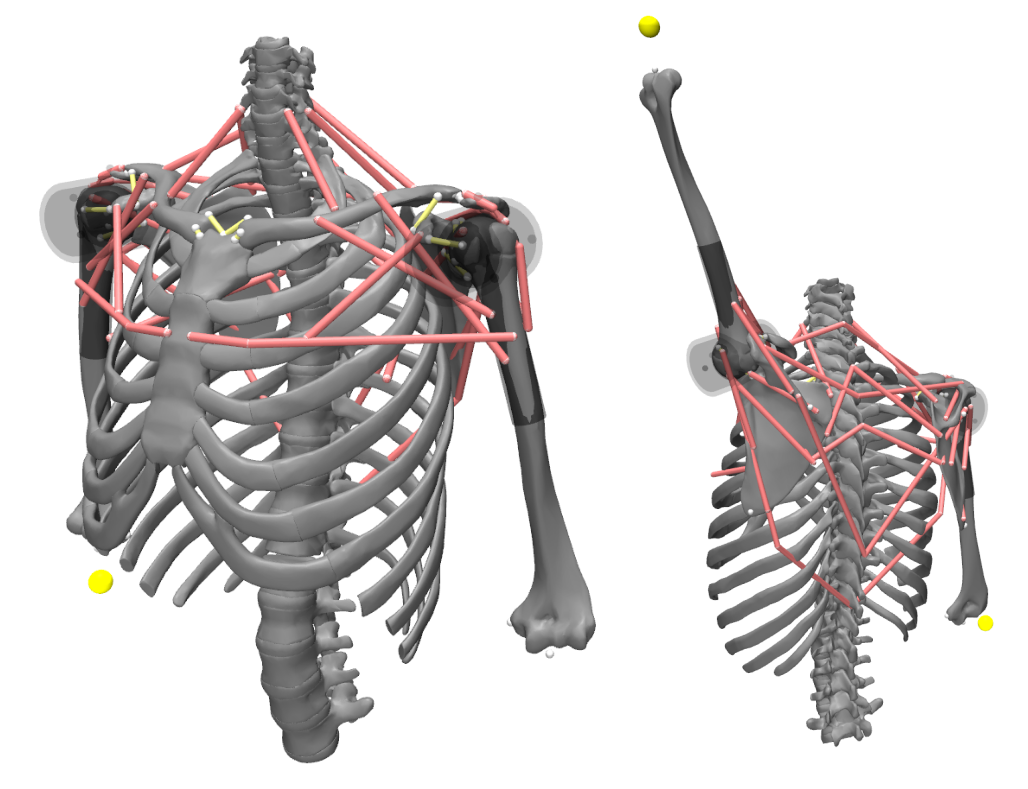}
  \caption{Overview of the proposed musculoskeletal model.}
  \label{fig:model_back}
\end{figure}
\section{Configuration of the musculoskeletal model of the shoulder complex}
\switchlanguage{
  In this study, it was necessary to use a simulator capable of simulating a musculoskeletal model with ligaments. 
  Since both muscles and ligaments are soft tissues, and muscles generate tension rather than torque like motors, 
  a simulator that can accurately represent these characteristics was required. 
  Additionally, the simulation speed was important when conducting robot simulations.

  Based on these considerations, this study utilized MuJoCo as the simulator \cite{todorov2012mujoco}.
  MuJoCo is a general-purpose physics engine intended to facilitate research and development in robotics, biomechanics, computer graphics and animation, machine learning, and other areas requiring fast and accurate simulation of articulated structures interacting with their environment.
  It can handle models containing tendons and muscles for simulations of tendon-driven robots such as musculoskeletal humanoids.
  Its simulation speed and accuracy are known to be superior to other mechanics simulators \cite{erez2015simulation}.
  Based on the above reasons, MuJoCo was employed in this study.

  The appearance of the musculoskeletal model of the shoulder complex created in this study is shown in \figref{fig:model_back}.
}
{
  本研究では、シミュレーターとして靭帯を伴う筋骨格モデルのシミュレーションを行うことができるものを用いる必要があった。
  筋肉と靭帯はいずれも軟部組織であり、さらに筋肉はモーターと違ってトルクではなく張力を発揮するため、
  これらを表現することができるシミュレータを用いる必要があった。
  また、ロボットのシミュレーションを行うにあたって、そのシミュレーション速度も重要であった。

  以上を踏まえ、本研究ではシミュレータにMuJoCo を用いた\cite{todorov2012mujoco}。
  MuJoCo は、
  ロボット工学、生体力学、CG・アニメーション、機械学習、および環境と
  相互作用する多関節構造の高速かつ正確なシミュレーションを必要とするその他の分野
  における研究開発を促進することを目的とした汎用物理エンジンである。
  筋骨格ヒューマノイドなどの腱駆動型ロボットのシミュレーション用に、腱や筋肉を含むモデルを扱うことができる。
  そのシミュレーション速度と精度は、他の力学シミュレータと比較しても優れていることが知られている\cite{erez2015simulation}。
  以上のような理由を踏まえて、本研究ではMuJoCo を採用した。

  本研究で作成した肩複合体の筋骨格モデルの外観を\figref{fig:model_back}に示す。
}
\switchlanguage{
  With the goal of creating a simulation model of the human shoulder complex with soft tissue ligaments and the same number of muscles as a human, this model was composed of three elements: skeleton, ligaments, and muscles.
  In this chapter, each of these elements is explained.
}
{
  人間の肩複合体のシミュレーションモデルに軟部組織である靭帯を有し、
  かつ人間と同等の数の筋肉を備えたモデルを作るという目的のもと、
  本モデルは骨格・靭帯・筋肉の3つの要素によって構成した。
  本章では、それぞれの要素について説明する。
}

\subsection{Linked structure of the skeleton}
\switchlanguage{
  This model was constructed by the bones that make up the shoulder complex: thorax, spine, scapula, clavicle, and humerus.
  Of these, the thorax and spine, which are included in the torso, are fixed to the world coordinate system, while the scapula, clavicle, and humerus, which are involved in shoulder motion, are connected by using the structural joints, the sternoclavicular joint, acromioclavicular joint, and scapulohumeral joint, as spherical counterpart joints.

  The links that compose this model were made using the skeletal mesh files provided by BodyParts3D\cite{mitsuhashi2009bodyparts3d}.
  BodyParts3D is a database developed by the Life Science Database Integration Project that describes the location and shape of human body parts (organs and organs) indicated by anatomical terms in a three-dimensional human body model.
  In addition to mesh files for the skeleton, it also provides mesh files for muscles and major organs, which were adopted because they could be used for judging skeletal hits and setting muscle attachment points.

  We attempted to construct the degrees of freedom of each joint based on bone-to-bone contact, but because it was difficult for MuJoCo to determine the concave surface contact, we modeled the joints as spherical counterpart joints.
  Therefore, this model could not reproduce phenomena caused by translational movement such as shoulder dislocation.
}
{
  本モデルは、肩複合体を構成する骨である胸郭・脊椎・肩甲骨・鎖骨・上腕骨によって構成した。
  このうち、胴体に含まれる胸郭と脊椎は世界座標系に固定し、
  肩の動きに関わる肩甲骨・鎖骨・上腕骨は、
  構造的関節である胸鎖関節、肩鎖関節、肩甲上腕関節を球面対偶関節とすることで繋いだ。

  このモデルを構成するリンクには、BodyParts3D\cite{mitsuhashi2009bodyparts3d} が提供する骨格のメッシュファイルを用いた。
  BodyParts3D は、ライフサイエンスデータベース統合推進事業にて開発された、
  解剖学用語が示す人体の部品（臓器・器官）の位置と形状を３次元人体モデルで記述したデータベースである
  。
  骨格のメッシュファイルの他、筋肉や主要な臓器のメッシュファイルも提供しており、
  骨格の当たり判定や筋肉の付着点の設定に利用できると考え採用した。

  各関節の自由度を骨同士の接触に基づき構成することを試みたが、
  MuJoCo では凹面の当たり判定が困難であったため、球面対偶関節としてモデル化した。
  そのため、このモデルでは肩の脱臼など並進方向の移動による現象は再現できていない。
}

\subsection{Representation and placement of ligaments}
\begin{figure}[t]
  \begin{minipage}[b]{1\linewidth}
    \centering
    \includegraphics[width=1\linewidth]{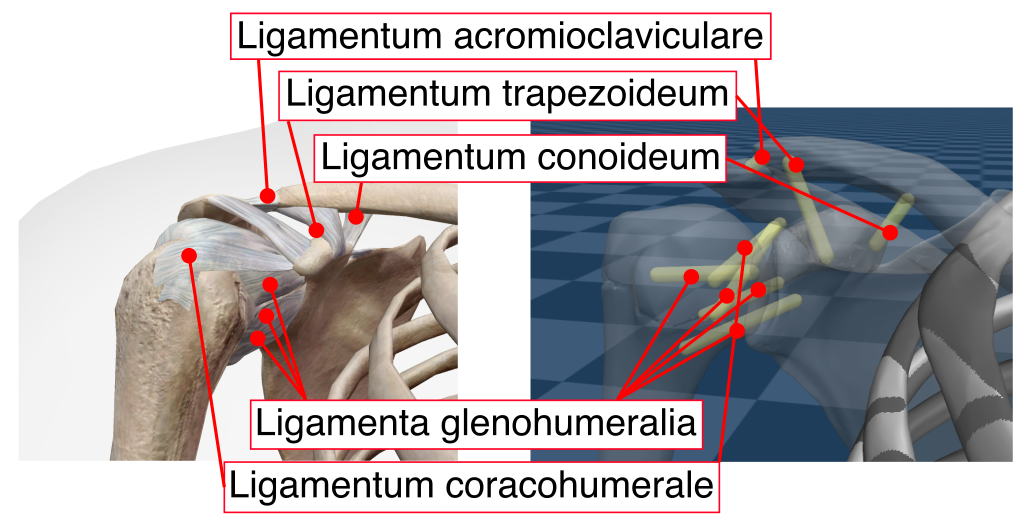}
    \subcaption{Ligaments. The yellow wire on right figure is the ligament.}
    \label{subfig:ligaments}
  \end{minipage}
  \begin{minipage}[b]{1\linewidth}
    \centering
    \includegraphics[width=1\linewidth]{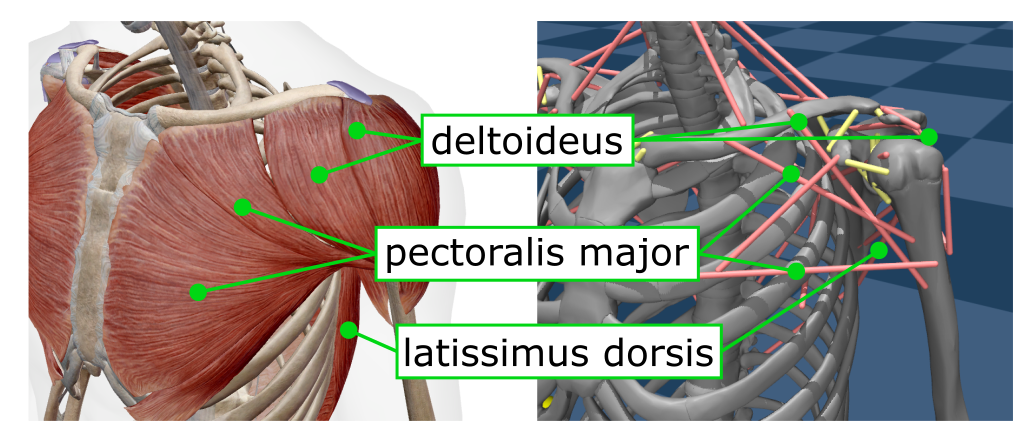}
    \subcaption{Muscles. The red wire on right figure is the muscle.}
    \label{subfig:muscles}
  \end{minipage}
  \centering
  \caption{Overview of the comparison between human and musculoskeletal models.
    The left figure shows the human from Visible Body ,
    and the right figure shows the musculoskeletal model developed in this study.}
  \label{fig:ligaments_muscles}
\end{figure}

\switchlanguage{
  Human bones are connected to each other by ligaments.
  Ligaments are connective tissues that bind bone to bone and are elastic biological soft tissues.
  It is mainly composed of collagen and has the role of maintaining the relative positional relationship between bones and joint stability through its elasticity.
  In this study, ligaments were represented as elastic wires.
  The elastic wires were placed as shown in \figref{subfig:ligaments}, based on the ligament and skeletal diagrams in Visible Body \cite{VisibleBody}.
  A list of the placed ligaments and the bones at both ends to which each ligament attaches is shown in \tabref{tab:ligaments}.
  MuJoCo's tendon element was used as the elastic wire.
}
{
  人間の骨は、靭帯によって結合されている。
  靭帯は、骨と骨結合する結合組織であり、弾性を持った生体軟組織である。
  主にコラーゲンなどにより構成され、弾性を持つことによって、
  骨同士の相対的な位置関係を保ち、関節の安定性を保つ役割を持つ。
  本研究では靭帯を弾性を持ったワイヤとして表現した。
  弾性を持ったワイヤを、\figref{subfig:ligaments}に示すように、
  Visible Body \cite{VisibleBody}の靭帯と骨格の図をもとに配置した。
  配置した靭帯の一覧と、それぞれの靱帯が付着する両端の骨を\tabref{tab:ligaments}に示す。
  尚、弾性を持ったワイヤとしてMuJoCoのtendon要素を用いた。
}
\begin{table}[t]
  \centering
  \caption{List of ligaments in the musculoskeletal model.
    The ``start'' and ``stop'' are bones with end points of ligaments.}
  \label{tab:ligaments}
  \begin{tabular}{lll} \hline
    name                            & start     & stop     \\ \hline
    Lig. interclaviculare           & manubrium & clavicle \\
    Lig. stemoclaviculare antarius  & manubrium & clavicle \\
    Lig. stemoclaviculare posterius & manubrium & clavicle \\
    Lig. acromioclaviculare         & clavicle  & scapula  \\
    Lig. trapezoideum               & clavicle  & scapula  \\
    Lig. coracohumerale             & clavicle  & scapula  \\
    Ligamenta glenohumeralia        & scapula   & humerus  \\ \hline
  \end{tabular}
\end{table}
\switchlanguage{
  In Visible Body, the ligaments and skeleton are shown in 3D, but their coordinates, etc., are not specifically indicated, so the coordinates of the end points of each ligament were determined visually, referring to the diagram of the ligaments in Visible Body.
  Although some ligaments do not constrain the positional relationship between bones, such as the transverse suprascapular ligament that forms the suprascapular foramen where the nerve passes, such ligaments were omitted in this model, and only ligaments that connect the different bones were placed.
  Since the glenohumeral ligament surrounds the humeral head of the scapulohumeral joint, it was represented by creating a sphere of the same shape as the humeral head and wrapping it around it.
  In this model, each joint was modeled as a spherical counterpart joint, so these ligaments were expected primarily to represent a range of motion equivalent to that of a human and to represent joint suppleness.
}
{
  Visible Body では、靭帯や骨格の図が3次元で表示されているが、
  その座標などが具体的に示されているわけではないため、
  Visible Body の靭帯の図を参考にしながら、目視で各靭帯の端点の座標を決定した。
  また、靭帯には、神経が通る肩甲上孔を形成する役割の上肩甲横靭帯など、
  骨同士の位置関係を拘束する役割ではないものも存在するが、
  本モデルでは、そのような人体は割愛し、異なる骨の間を繫ぐ靭帯のみを配置した。
  関節上腕靱帯は肩甲上腕関節の上腕骨頭を囲むように付いているため、
  上腕骨頭と同じ形状の球を作成し、その球に巻き付かせることによって表現した。
  本モデルにおいては、各関節は球面対偶関節としてモデル化されているため、
  これらの靭帯は、主に人間と同等の関節可動域を表現することや、関節のしなやかさを表現することが期待された。
}

\subsection{Representation and placement of muscles}
\switchlanguage{
  Human skeletal muscles can be regarded as actuators between bones that generate tension.
  Bones and muscles are connected by tendons.
  Muscles are activated by electrical signals from nerves to generate tension, and mathematical models such as cross-bridge models \cite{eisenberg1980cross} and Hill-type models \cite{hill1974theoretical} and other mathematical models exist to describe this behavior.
  The former model is based on the basic structure inside the muscle, but there are many parameters that must be identified, and many simulators use Hill-type models \cite{millard2013musclemodel}.

  In this study, Hill-type models were used to represent muscles.
  The muscles involved in driving the shoulder complex were placed as shown in \figref{subfig:muscles}.
  A list of the placed muscles and the origin and stop of each muscle attachment is shown in \tabref{tab:muscles}.
  The tendon and muscle elements of MuJoCo were used as muscles.
}
{
  人間の骨格筋は、骨と骨の間に、張力を発生させるアクチュエータとみなせる。
  骨と筋肉は腱によって結合している。
  筋肉は神経から電気的な信号を受けることにより活性化し、張力を発生させており、
  その振る舞いの記述のために、cross-bridge models\cite{eisenberg1980cross}やHill-type models\cite{hill1974theoretical}などの数理モデルが存在する。
  前者は、筋肉の内部の基本構造に基づいたモデルであるが、同定しなければいけないパラメータが多く、
  多くのシミュレータではHill-type modelsが用いられる\cite{millard2013musclemodel}。

  本研究では、筋肉の表現にHill-type modelsを用いた。
  \figref{subfig:muscles}に示すように、肩複合体の駆動にかかわる筋肉を配置した。
  配置した筋肉の一覧と、それぞれの筋肉が付着する起始と停止を\tabref{tab:muscles}に示す。
  尚、筋肉としてMuJoCoのtendon要素とmuscle要素を用いた。
}
\begin{table}[t]
  \centering
  \caption{List of muscles in the musculoskeletal model.
    The ``origin'' and ``insertion'' are bones with end points of muscles.}
  \label{tab:muscles}
  \begin{tabular}{lll} \hline
    name                               & origin   & insertion \\ \hline
    pectoralis major                   & thorax   & humerus   \\
    pectoralis minor                   & thorax   & scapula   \\
    subscapularis                      & scapula  & humerus   \\
    infraspinatus                      & scapula  & humerus   \\
    teres major                        & scapula  & humerus   \\
    teres minor                        & scapula  & humerus   \\
    serratus anterior                  & thorax   & scapula   \\
    pectoralis major pars clavicularis & clavicle & humerus   \\
    latissimus dorsi                   & thorax   & humerus   \\
    deltoideus pars spinalis           & scapula  & humerus   \\
    deltoideus pars acriminalis        & scapula  & humerus   \\
    deltoideus pars clavicularis       & clavicle & humerus   \\
    supraspinatus                      & scapula  & humerus   \\
    platysma                           & thorax   & clavicle  \\
    levator scaplulae                  & thorax   & scapula   \\
    rhomboideus minor                  & thorax   & scapula   \\
    rhomboideus major                  & thorax   & scapula   \\
    descending part of trapezius       & thorax   & clavicle  \\
    transverse part of trapezius       & thorax   & scapula   \\
    ascending part of trapezius        & thorax   & scapula   \\ \hline
  \end{tabular}
\end{table}

\switchlanguage{
  Humans have approximately 640 muscles.
  Since this study focuses only on the shoulder complex, 40 muscles that drive the scapula, humerus, and clavicle were selected.
  The way the muscles are classified varies somewhat in the literature, but in this study we followed the mesh file of the muscles contained in the BodyParts3D.
  Therefore, some muscles, such as the trapezius, were divided into multiple regions, and independent muscles were placed for each region in this model.
  The endpoints of each muscles were placed at the starting and stopping points of the muscles, respectively, referring to the mesh files of the muscles in BodyParts3D.
  Since both the starting and stopping points are not simple points but are in plane contact with the bone, in this model, they were placed at coordinates near the center of the plane by referring to the mesh file of the muscles.
  In addition, some muscles, such as the deltoid and pectoralis major, generate torque on the bone by wrapping around it.
  A cylinder was placed in the same position as the humerus, and muscles were wrapped around this cylinder to represent the wrapping of the muscles around the bone.
}
{
  人間には約640の筋肉が存在するが、本研究では肩複合体にのみ注目するため、
  肩甲骨・上腕骨・鎖骨を駆動する40筋を選択した。
  筋肉の区分の仕方は、文献によって多少異なるが、
  本研究ではBodyParts3D に含まれる筋肉のメッシュファイルに従った。
  そのため、僧帽筋など、一部の筋肉は複数の部位に分割されており、
  本モデルでは、それぞれの部位に対して独立した筋肉を配置した。

  各筋肉の端点は、BodyParts3D の筋肉のメッシュファイルを参照し、
  その起始部と停止部それぞれに配置した。
  起始部、停止部はいずれも単純な点ではなく、骨と面で接している部分であるため、
  本モデルでは、筋肉のメッシュファイルを参照し、その面の中心付近の座標に配置した。

  また、三角筋や大胸筋などの一部の筋肉は、骨に巻き付くことにより骨にトルクを発生させる。
  上腕骨と同じ位置に円筒を配置し、これに筋肉を巻き付けることで、筋肉の骨への巻き付きを表現した。
}

\switchlanguage{
  In the muscle element of MuJoCo, there exists a parameter that sets the maximum muscle force.
  It is known that there is a proportional relationship between the maximum muscle force and physiological cross-sectional area of human muscles such that \equref{eq:force_pcsa}\cite{pcsa_force}.
}
{
  MuJoCo のmuscle要素には、最大筋力を設定するパラメータが存在する。
  人間の筋肉の最大筋力と生理学的断面積には、
  \equref{eq:force_pcsa}のような比例関係
  があることが知られている\cite{pcsa_force}。
}
\begin{align}
  F           & = k_{stress} \cdot \text{PCSA} \label{eq:force_pcsa}  \\
  \text{PCSA} & = \frac{V}{l} \cdot \cos\alpha \label{eq:pcsa_volume} \\
  V           & = \frac{m}{\rho} \label{eq:volume_mass}
\end{align}
\switchlanguage{
  $F$ is the maximum muscle force, $k_{stress}$ is the proportionality constant, and PCSA is the physiological cross-sectional area of the muscle.
  The physiological cross-sectional area is defined by \equref{eq:pcsa_volume}, where $V$ is muscle volume, $l$ is muscle length, $\alpha$ is feather angle, $m$ is muscle mass, and $\rho$ is muscle density.
  This allows us to estimate the maximum muscle force if the muscle volume and length are known.
  In this model, the maximum muscle force was estimated by defining the muscle volume $V$ by the volume of the mesh file provided by BodyParts3D and the muscle length $l$ by the length of the tendon element on the model.
  The proportionality constant $k_{stress}$ was set to the literature value \SI{59}{Pa} \cite{HALE2011650}.
}
{
  ただし、$F$は最大筋力、$k_{stress}$は比例定数、PCSAは筋肉の生理学的断面積である。
  生理学的断面積は、\equref{eq:pcsa_volume}によって定義され、
  $V$は筋体積、$l$は筋肉の長さ、$\alpha$は羽状角、$m$は筋肉の質量、$\rho$は筋肉の密度である。
  これにより、筋肉の体積と長さが分かれば、最大筋力を推定することができる。
  本モデルでは、筋体積$V$をBodyParts3D によって提供されるメッシュファイルの体積によって定義し、
  筋肉の長さ$l$をモデル上のtendon要素の長さによって定義することで、最大筋力を推定した。
  比例定数$k_{stress}$は文献値\SI{59}{Pa}とした\cite{HALE2011650}。
}

\section{Position control by model predictive control}
\begin{figure}[t]
  \centering
  \includegraphics[width=1\linewidth]{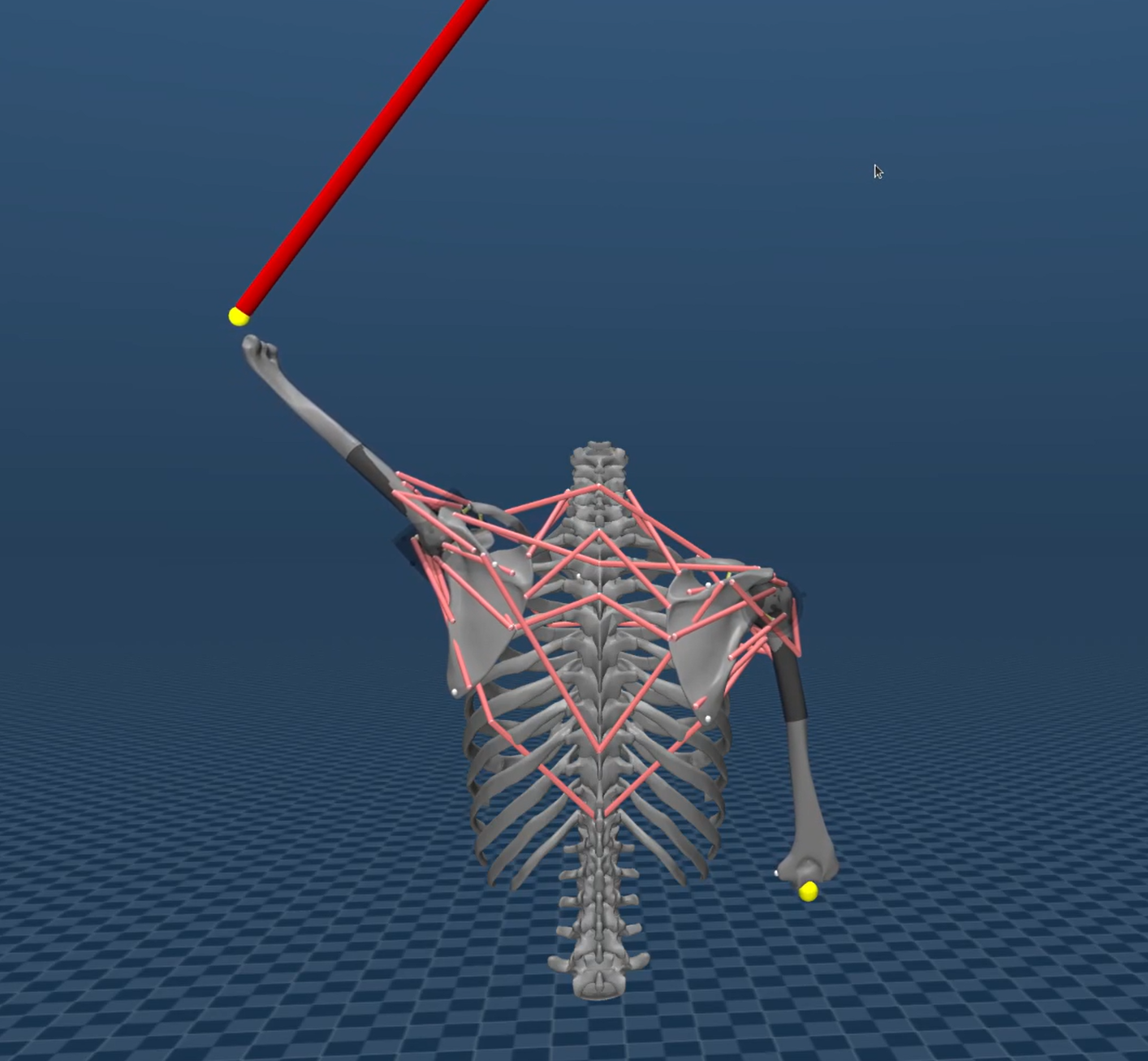}
  \caption{Position control of the proposed musculoskeletal model with MPC.
    The two yellow balls are the target positions.
    The muscles are controlled to bring the tip of the humerus closer to the target position.}
  \label{fig:mjpc}
\end{figure}
\switchlanguage{
  The model developed had 40 muscles, making it a multi-input system.
  To see if this could be controlled, positional control of the elbow tip was performed.
  Because human muscles are arranged in an antagonistic manner, they can maintain their position even if extra muscular force is exerted between antagonistic muscles.
  However, exerting extra muscular force is thought to promote muscle fatigue, which is considered to be detached from actual human movement.
  Therefore, in this study, optimal control was employed to suppress unnecessary muscle fatigue.
  As a modeling method for muscle fatigue, there exists a model focusing on motor units proposed by Liu et al \cite{liu2002dynamical}.
  According to this model, muscle fatigue is calculated by the differential equation between muscle fatigue and the signal input to the muscle; the greater the signal to the muscle, the more muscle fatigue accumulates.
  Therefore, the objective of reducing muscle fatigue was substituted by simply decreasing the signal sent to the muscle.

  In this study, model predictive control was employed to achieve optimal control in a multi-input system.
  As an experimental environment, we used the software MuJoCo MPC, which is capable of model predictive control on the CPU\cite{howell2022mjpc}.
  MuJoCo MPC is software for online model predictive control on MuJoCo. By setting the model and cost function, control can be performed on MuJoCo through model predictive control.
  The target position was set by a yellow ball, and by moving this, it was checked whether the elbow tip would follow.
  In MuJoCo MPC, optimal control is performed such that the cost function is minimized.
  The cost function $J$ is defined as the sum of the cost with respect to the position $\bs{p}$ and the cost with respect to the control input $\bs{u}$ with weights $w_i$ as shown in \equref{eq:cost_function}.
}
{
  開発したモデルは、40筋を有しており、多入力系となった。
  これを制御することができるかどうかを確認するために、肘先の位置制御を行った。
  人間の筋肉は拮抗するように配置されているため、
  拮抗する筋肉同士に余分な筋力を発揮させても、位置を保つことができる。
  しかし、余分な筋力を発揮することは、筋疲労を促進してしまい、
  実際の人間の動作とは離れてしまうと考えられる。
  そこで、本研究では、無駄な筋疲労を抑えるために、最適制御を採用した。

  筋疲労のモデル化手法として、Liuらが提案した運動単位に着目したモデルが存在する\cite{liu2002dynamical}。
  このモデルによると、筋疲労と筋肉への信号入力の間の微分方程式によって筋疲労が計算され、
  筋肉への信号が大きいほど筋疲労が溜まりやすくなる。
  そのため、本研究では筋疲労を抑えるという目的を、単純に筋肉に送る信号を小さくすることで代替した。

  以上より、本研究では多入力系における最適制御を行うため、モデル予測制御を採用した。
  実験環境として、MuJoCo 上でモデル予測制御を行うことのできるソフトウェアであるMuJoCo MPC を用いた\cite{howell2022mjpc}。
  MuJoCo MPC は、MuJoCo においてオンラインでモデル予測制御を行うためのソフトウェアであり、
  MuJoCo のモデルとコスト関数を設定することで、モデル予測制御によりMuJoCo 上で制御を行うことができる。

  目標位置を黄色い球によって設定し、これを動かすことで、肘先が追従するかどうかを確認した。
  MJPCでは、コスト関数を最小化するような最適制御が行われる。
  コスト関数$J$は\equref{eq:cost_function}に示すように位置$\bs{p}$に関するコストと制御入力$\bs{u}$に関するコストの重み$w_i$付きの和として定義した。
}
\begin{align}
  J   & = \sum_{i} w_i \cdot r_i   \label{eq:cost_function}                                          \\
  r_1 & = ||\bs{p}_{\mathrm{target}} - \bs{p}_{\mathrm{endeffector}}|| \label{eq:mjpc_cost_position} \\
  r_2 & = ||\bs{u}||^2 \label{eq:mjpc_cost_control}
\end{align}
\switchlanguage{
\equref{eq:mjpc_cost_position} is a constraint on position, with the distance between the end-effector and the target position as the cost.
$\bs{p}_{\mathrm{target}}$ is the target position and $\bs{p}_{\mathrm{endeffector}}$ is the end-effector position.
In this study, the end-effector position was set at the tip of the humerus.
\equref{eq:mjpc_cost_control} is a constraint on the control input, and the cost is the sum of squares of the vector $\bs{u}\in [0,1]^{40}$ that lines up the signals to the 40 muscles in the model.
}
{
\equref{eq:mjpc_cost_position}は位置に関する制約であり、
エンドエフェクタと目標位置の距離をコストとした。
$\bs{p}_{\mathrm{target}}$は、目標位置であり、
$\bs{p}_{\mathrm{endeffector}}$は、エンドエフェクタの位置である。
本研究ではエンドエフェクタの位置を上腕骨の先端に設定した。
\equref{eq:mjpc_cost_control}は制御入力に関する制約であり、
モデルに含まれる40本の筋肉への信号を並べたベクトル$\bs{u}\in [0,1]^{40}$の2乗和をコストとした。
}
\section{Experiments with position control}
\subsection{Movement experiment by position control of elbow tip}
\begin{figure}[t]
  \begin{minipage}[b]{1\linewidth}
    \centering
    \includegraphics[width=1\linewidth]{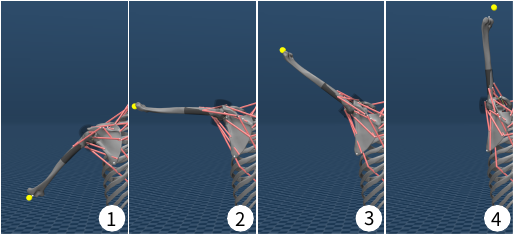}
    \subcaption{Abduction}
    \label{subfig:abdaction}
  \end{minipage}
  \begin{minipage}[b]{1\linewidth}
    \centering
    \includegraphics[width=1\linewidth]{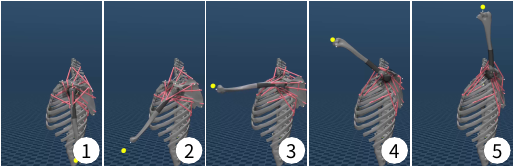}
    \subcaption{Forward flexion}
    \label{subfig:flexion}
  \end{minipage}
  \begin{minipage}[b]{1\linewidth}
    \centering
    \includegraphics[width=1\linewidth]{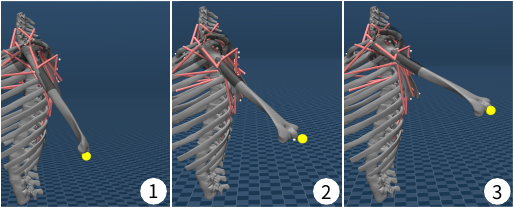}
    \subcaption{Backward extension}
    \label{subfig:extension}
  \end{minipage}
  \centering
  \caption{Elevation movements with MPC. As in humans, approximately 180 degrees of abduction, 180 degrees of anterior flexion, and 50 degrees of posterior flexion can be observed.}
  \label{fig:flexion}
\end{figure}
\switchlanguage{
  We verified that the implemented elbow tip position control works correctly.
  It was confirmed that the elbow tip followed the yellow ball, which was set as the target position, by moving it manually.
  This is shown in \figref{fig:mjpc}.
  When the model's arm is moved by position control, its range of motion should be equivalent to that of a human.
  The human shoulder can be raised approximately 180 degrees forward, 180 degrees laterally, and 50 degrees backward \cite{kapandji2007physiology}.
  Therefore, we checked to what extent the shoulder could be raised in this model anteriorly, laterally, and posteriorly, respectively.
  For each direction, the shoulder is shown in \figref{fig:flexion} when it is elevated.
  Here, the target position was set for each angle of elevation, and the state was observed after the shoulder was elevated and a short wait was made for the position to stabilize.
  \figref{subfig:abdaction} shows the state when the arm is raised laterally, \figref{subfig:flexion} shows the state when the arm is raised forward, and \figref{subfig:extension} shows the state when the arm is raised backward.
  This confirms that the range of motion in both directions is comparable to that of a human.
}
{
  実装した肘先の位置制御が正しく動作するか試した。
  目標位置とした黄色い球を手動で動かし、これに肘先が追従することを確認した。
  その時の様子を\figref{fig:mjpc}に示す。
  また、位置制御によってモデルの腕を動かしたとき、
  その可動域は人間と同等になっているべきである。
  人間の肩は、前方に約180度、側方に約180度、後方に約50度挙上することができる\cite{kapandji1971physiology}。
  よって、本モデルにおいて、肩を前方、側方、後方にそれぞれどこまで挙上できるかを確認した。
  各方向について、肩を挙上したときの様子を\figref{fig:flexion}に示す。
  ここでは、挙上角ごとに目標位置を設定し、
  肩を挙上してから位置が安定するまで少し待った後の状態を観察した。
  \figref{subfig:abdaction}には腕を側方に挙上した時の状態を、
  \figref{subfig:flexion}には腕を前方に挙上した時の状態を、
  \figref{subfig:extension}には腕を後方に挙上した時の状態をそれぞれ示している。
  これによって、いずれの方向についても、人間と同程度の可動域を持つことが確認できた。
}
\subsection{Raising movement experiment with and without ligaments}
\begin{figure}[t]
  \begin{minipage}[b]{0.48\linewidth}
    \centering
    \includegraphics*[width=1\linewidth,trim=656px 100px 800px 91px]{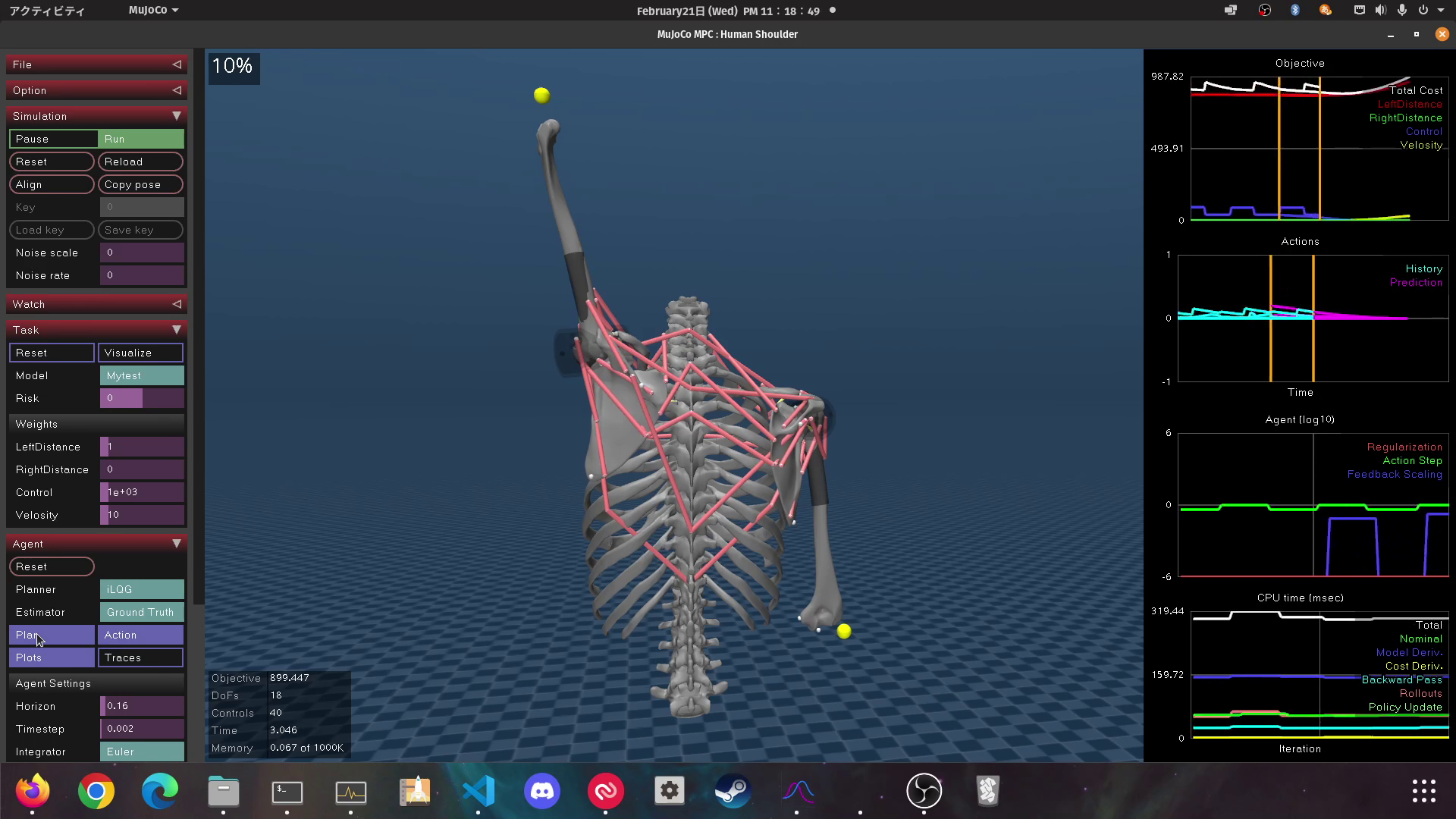}
    \subcaption{With ligaments}
  \end{minipage}
  \begin{minipage}[b]{0.48\linewidth}
    \centering
    \includegraphics*[width=1\linewidth,trim=656px 100px 800px 91px]{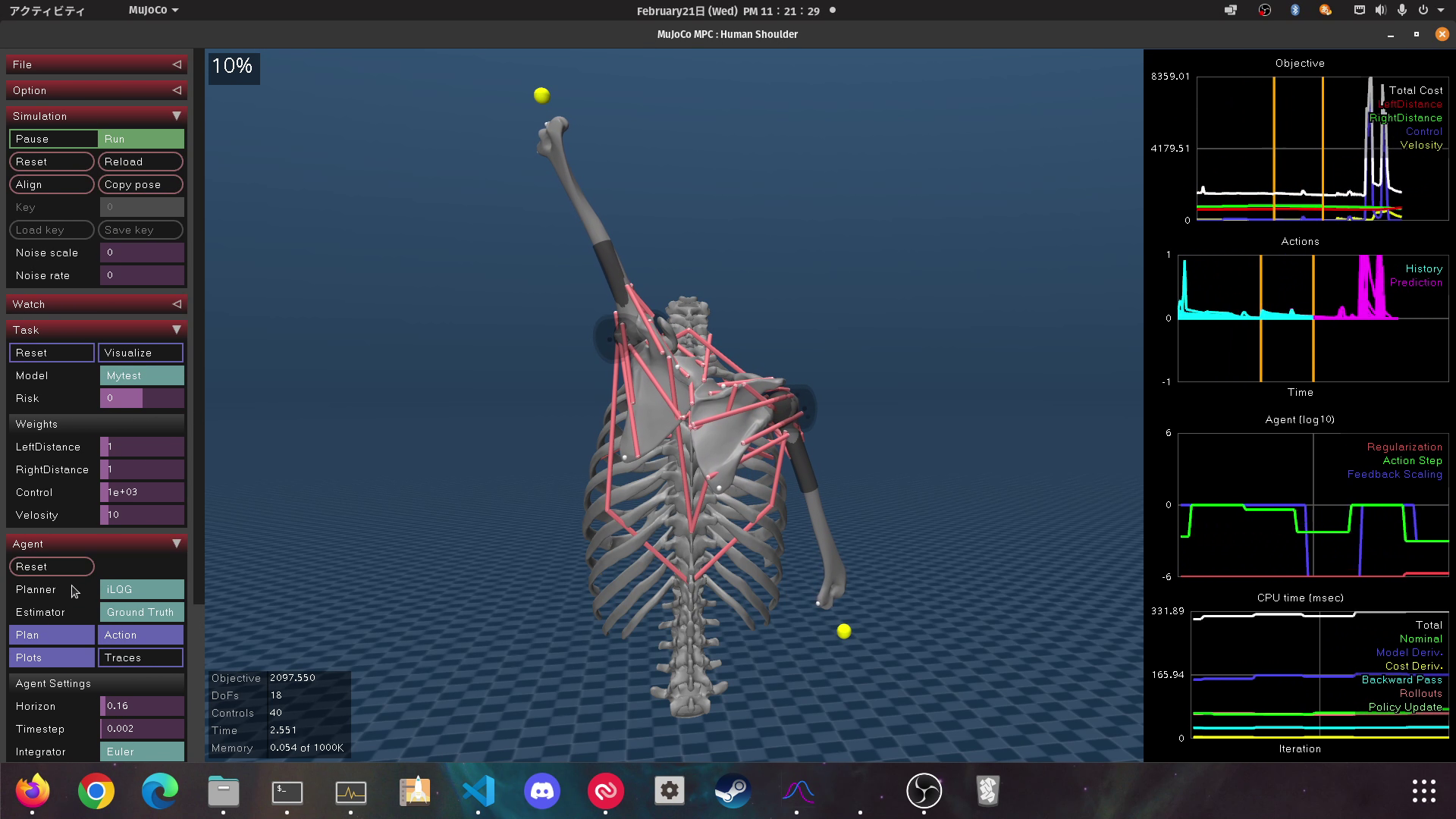}
    \subcaption{Without ligaments}
  \end{minipage}
  \centering
  \caption{Comparison of the elevation of the upper arm with and without ligaments. The position of the scapula is closer to the spine when the ligaments are absent.}
  \label{fig:with_without_ligaments}
\end{figure}

\switchlanguage{
  To see how the joint range of motion limitation of the ligament affects the raising of the upper arm, an experiment was conducted in which the elbow was raised with and without the ligament, respectively.
  By setting the target position above the left shoulder, the model was placed in an elevated posture.
  After the elbow was raised, the model waited until the posture became stable, and then was observed.
  The situation is shown in \figref{fig:with_without_ligaments}.
  Compared to when the ligaments were present, when the ligaments were absent, the position of the scapula was closer to the spine.
  The elimination of the ligaments has increased the range of motion of the sternoclavicular and acromioclavicular joints, allowing the shoulder to be positioned closer to the center of the body.
  This is thought to be due to the fact that the weight of the shoulder is now more easily deposited on the body.
}
{
  靭帯の関節可動域制限が、上腕の挙上にどのような影響を与えるかを確認するために、
  靭帯がある状態と無い状態それぞれで肘を挙上する実験を行った。
  上腕を挙上させるような位置に目標位置を設定することで、挙上姿勢をとらせた。
  挙上してから位置が安定するまで待った後の状態を観察した。
  その様子を\figref{fig:with_without_ligaments}に示す。
  靭帯がある時と比較して、靭帯がない時は、肩甲骨の位置が脊椎に近づいている。
  靭帯を無くしたことで、胸鎖関節や肩鎖関節の可動域が広がり、
  肩の位置をより身体の中心に近づけることができるようになった。
  これにより、肩の重量を身体に預けやすくなったためと考えられる。
}

\begin{figure}[t]
  \begin{minipage}[b]{1\linewidth}
    \centering
    \includegraphics[width=1\linewidth]{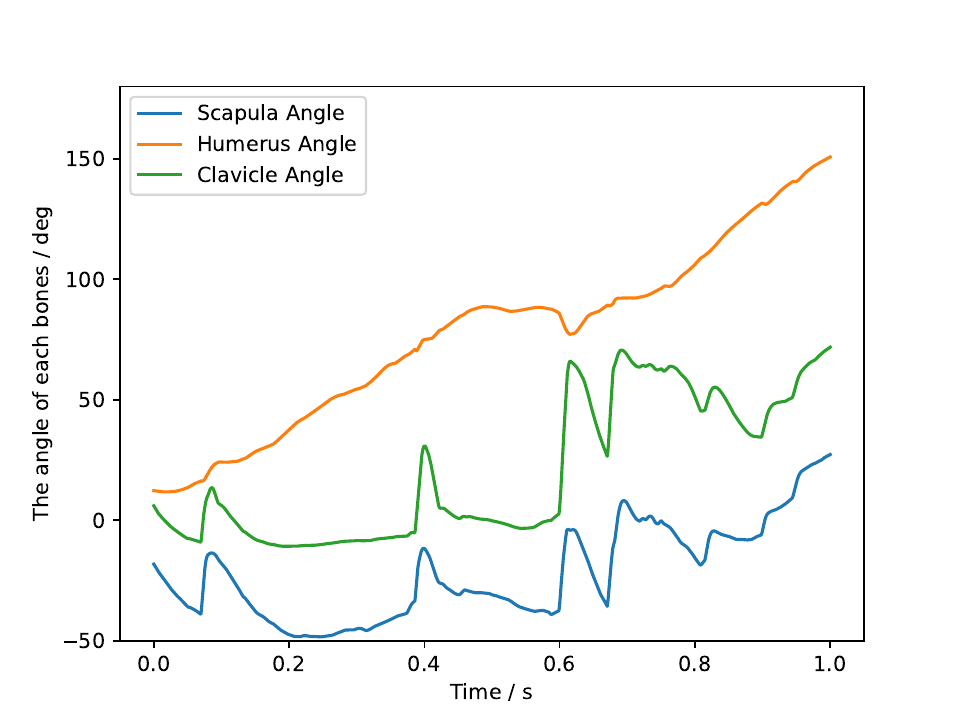}
    \subcaption{Without ligaments}
  \end{minipage}
  \begin{minipage}[b]{1\linewidth}
    \centering
    \includegraphics[width=1\linewidth]{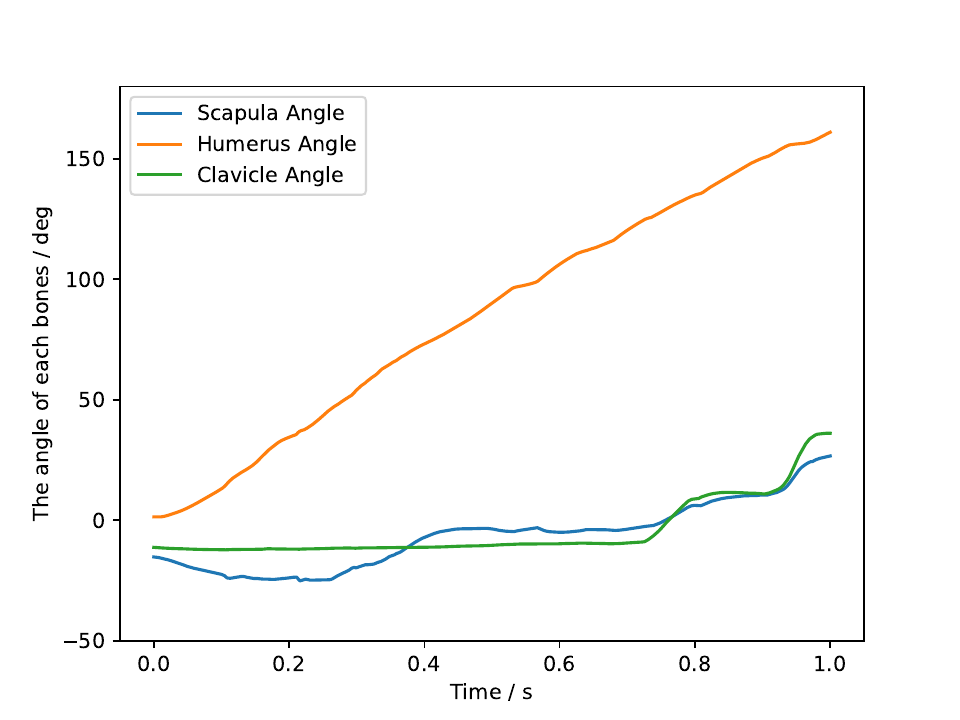}
    \subcaption{With ligaments}
  \end{minipage}
  \centering
  \caption{Angle of rotation of scapula, clavicle, and humerus during raising motion.}
  \label{fig:stability}
\end{figure}

\switchlanguage{
  The angles of the clavicle, scapula, and humerus during the raising motion are also shown in \figref{fig:stability}.
  This shows that the angular changes of the scapula and clavicle are more oscillatory when the ligaments are absent than when they are present, indicating that the lack of range-of-motion restriction by the ligaments makes that movement unstable.
}
{
  また、挙上動作中の鎖骨・肩甲骨・上腕骨の角度を\figref{fig:stability}に示す。
  これをみると、靭帯がある時と比較して、靭帯がない時は、肩甲骨と鎖骨の角度変化が振動的であり、
  靭帯による可動域制限がないことでその動作が不安定になっていることを示している。
}

\subsection{Raising movement experiment with estimated and averaged maximum muscle force}
\switchlanguage{
  In this model, the maximum muscle force of each muscle was estimated from muscle volume and length.
  In order to confirm how this maximum muscle force affects the upper arm raising motion, an experiment was conducted to raise the elbow when the maximum muscle force was estimated from the muscle volume and when it was averaged.
  Since maximal muscle force is proportional to muscle volume, averaging it corresponds to redistributing the muscle volumes of each muscle so that they are equal while maintaining the total muscle volume of each muscle.
  The maximum muscle force estimated from the muscle volume and the maximum muscle force when averaged are shown in \tabref{tab:maximum_muscle_forces}.
}
{
  本モデルでは、各筋肉の最大筋力を筋体積から推定した。
  この最大筋力が、上腕の挙上にどのような影響を与えるかを確認するために、
  最大筋力を筋体積から推定した時と平均化した時で肘を挙上する実験を行った。
  最大筋力は筋体積と比例するため、これを平均化するということは、
  各筋肉の筋体積の総量を変えずにそれぞれの筋体積が等しくなるよう再分配することに相当する。
  最大筋力を筋体積から推定した時と、平均化した時それぞれの最大筋力を\tabref{tab:maximum_muscle_forces}に示す。
}
\begin{table}[t]
  \centering
  \caption{Maximum muscle forces in the musculoskeletal model. EMF is estimated muscle force, and AMF is average muscle force. The underlined values are higher than the average muscle force.}
  \label{tab:maximum_muscle_forces}
  \begin{tabular}{lrr} \hline
    muscle                             & EMF / $\si{N}$     & AMF / $\si{N}$ \\ \hline \hline
    pectoralis major                   & \underline{445.0}  & 276.9          \\
    pectoralis minor                   & 23.0               & 276.9          \\
    subscapularis                      & \underline{417.0}  & 276.9          \\
    infraspinatus                      & 140.0              & 276.9          \\
    teres major                        & 98.5               & 276.9          \\
    teres minor                        & 79.6               & 276.9          \\
    serratus anterior                  & 40.0               & 276.9          \\
    pectoralis major pars clavicularis & 176.0              & 276.9          \\
    latissimus dorsi                   & 197.0              & 276.9          \\
    deltoideus pars spinalis           & 32.6               & 276.9          \\
    deltoideus pars acriminalis        & 32.4               & 276.9          \\
    deltoideus pars clavicularis       & 118.0              & 276.9          \\
    supraspinatus                      & 146.0              & 276.9          \\
    platysma                           & \underline{799.0}  & 276.9          \\
    levator scaplulae                  & 22.2               & 276.9          \\
    rhomboideus minor                  & 35.2               & 276.9          \\
    rhomboideus major                  & 30.0               & 276.9          \\
    descending part of trapezius       & \underline{2626.0} & 276.9          \\
    transverse part of trapezius       & 32.1               & 276.9          \\
    ascending part of trapezius        & 48.2               & 276.9          \\ \hline
    total                              & 5537.8             & 5537.8         \\ \hline
  \end{tabular}
\end{table}

\begin{figure}[t]
  \begin{minipage}[b]{1\linewidth}
    \centering
    \includegraphics[width=1\linewidth]{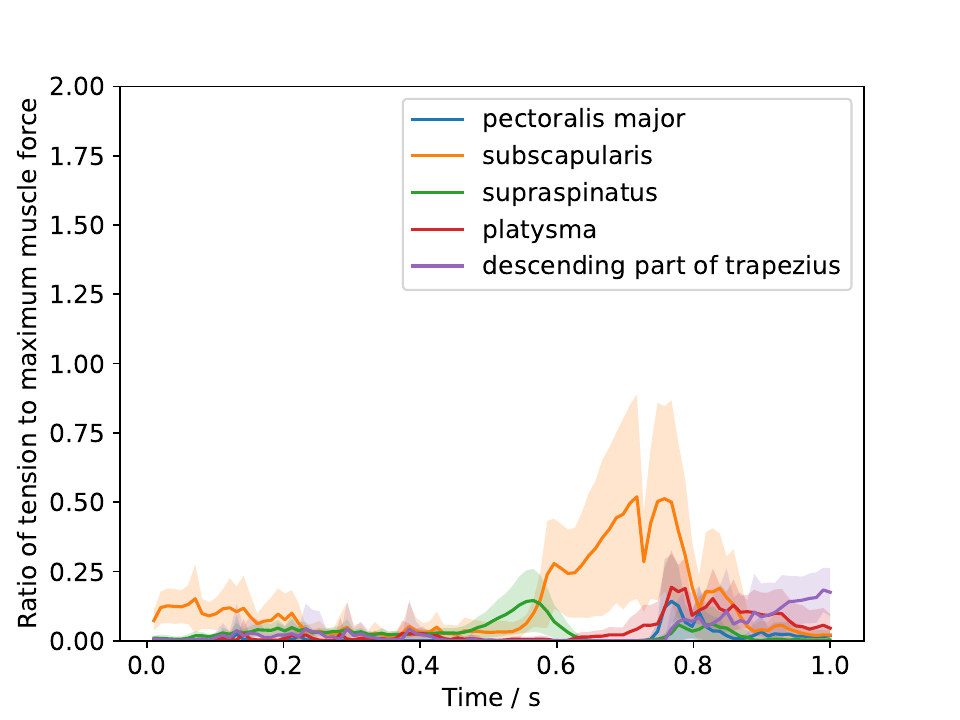}
    \subcaption{Estimated}
  \end{minipage}
  \begin{minipage}[b]{1\linewidth}
    \centering
    \includegraphics[width=1\linewidth]{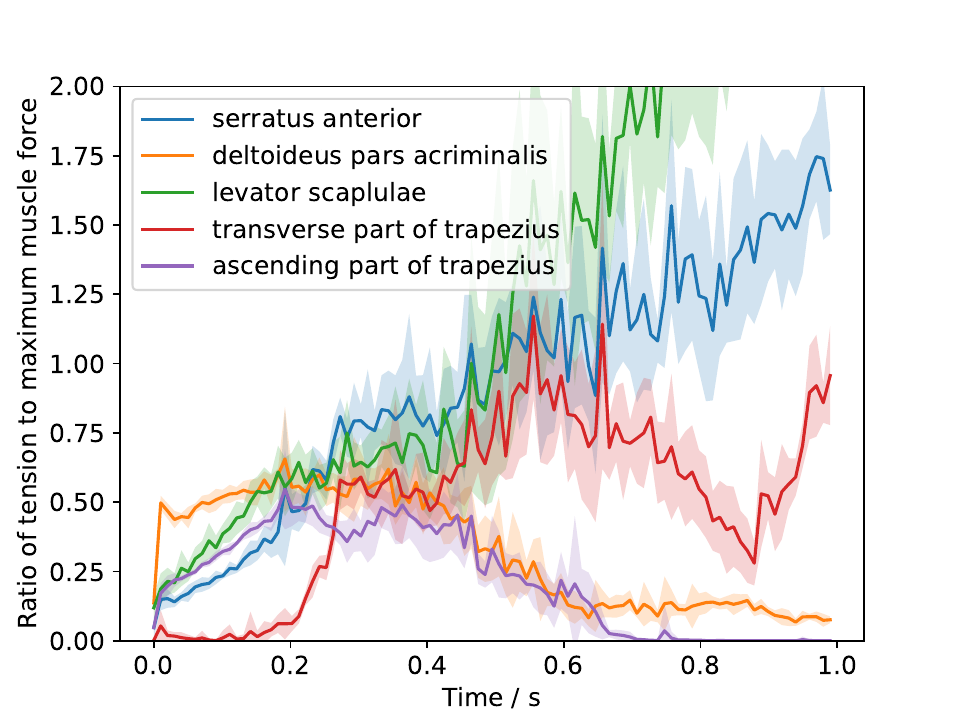}
    \subcaption{Averaged}
  \end{minipage}
  \centering
  \caption{Comparison of the transitions and distribution of muscle tension during the arm raising movement in the case of estimated and averaged muscle tension.
  For the mean $\mu$ and standard deviation $\sigma$ of the muscle tension at each time during the 8 trials,
  the change in $\mu$ is represented by a line,
  and the distribution is shown by filling in the $[\mu - \sigma,\mu + \sigma]$ area.}
  \label{fig:estimate_average_actuator}
\end{figure}

\switchlanguage{
  An experiment was conducted in which the target position was moved in a semicircle around the shoulder to raise the upper arm over a period of one second.
  Eight trials were made for each of the times when the maximum muscle force was estimated from muscle volume, etc. and when it was averaged, and the tension force applied to the muscle at this time was recorded.
  The ratio of the recorded tension divided by the maximum muscle force was calculated, and the average $\mu$ is shown in \figref{fig:estimate_average_actuator}.
  In addition, the standard deviation $\sigma$ at each time is shown by coloring the area included in $[\mu-\sigma,\mu+\sigma]$.
}
{
  目標位置を肩を中心とした半円上を動かすことで、1秒かけて上腕を挙上させる実験を行った。
  それぞれの最大筋力設定について、8回ずつ試行し、この時の筋肉にかかっている張力を記録した。
  記録した張力を最大筋力で割った比を求め、その平均$\mu$を\figref{fig:estimate_average_actuator}に示す。
  同時に、各時刻での標準偏差$\sigma$を、
  $[\mu-\sigma,\mu+\sigma]$に含まれる部分に色を付けることにより示す。
}
\switchlanguage{
  This shows that some muscles have a larger ratio of tension to maximum muscle force when averaged than when maximum muscle force is estimated from muscle volume and when averaged.
  In particular, the tension ratio is larger in muscles such as the serratus anterior, levator scapulae, and descending trapezius.
  For the serratus anterior and levator scapulae muscles, the tension ratio is greater than 1, indicating that the muscles are being pulled and stretched fully.

  The volume of human muscles differs from region to region, and this in turn causes the maximum muscle strength to differ.
  Larger muscles, i.e., those with greater maximal muscle strength, are considered to be those that require greater force in actual movement.
  By averaging the maximum muscle strength, the muscles that originally demanded greater strength would not be able to exert sufficient muscle strength, and in order to raise the arm in this state, a greater load would have been placed on other muscles.
}
{
  これを見ると、最大筋力を筋体積から推定した時と平均化した時で、
  平均化したときの方が最大筋力に対する張力の比が大きくなっている筋があることがわかる。
  特に、前鋸筋や肩甲挙筋、僧帽筋下降部等の筋肉において、張力比が大きくなっている。
  前鋸筋や肩甲挙筋については張力比が1よりも大きくなっており、これは筋肉が引っ張られて伸び切っていることを示している。

  人間の筋肉は部位によってその体積が異なり、これによって最大筋力も異なる。
  大きい筋肉、つまり最大筋力の大きい筋肉は実際の動作の中でも大きな力が要求される筋肉と考えられる。
  最大筋力を平均化したことで本来大きな力が要求される筋肉が十分な筋力を発揮できなくなり、
  その状態で腕を挙上させるために他の筋肉に大きな負荷がかかるようになったのだろう。
}

\begin{figure}[t]
  \begin{minipage}[b]{0.48\linewidth}
    \centering
    \includegraphics*[width=1\linewidth,trim=656px 400px 1000px 91px]{mjpc/estimate.png}
    \subcaption{Estimated}
  \end{minipage}
  \begin{minipage}[b]{0.48\linewidth}
    \centering
    \includegraphics*[width=1\linewidth,trim=656px 400px 1000px 91px]{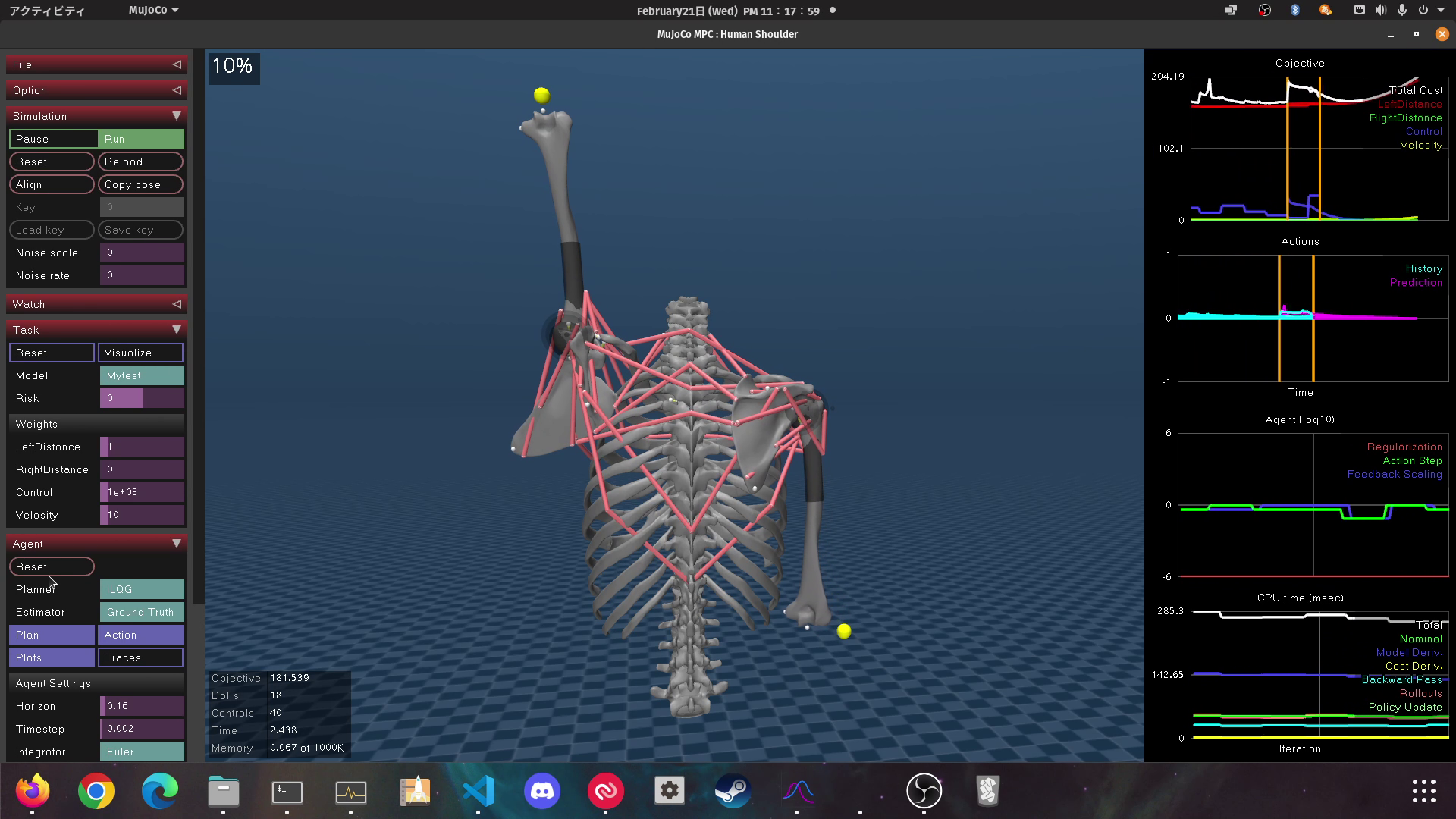}
    \subcaption{Average}
  \end{minipage}
  \centering
  \caption{Comparison of maximum muscle strength estimated from muscle volume and length and averaged. The scapula is further away from the spine in the averaged case than in the estimated case.}
  \label{fig:estimate_average}
\end{figure}

\switchlanguage{
  In addition, the image after the model's upper arm is raised and stabilized is shown in \figref{fig:estimate_average}.
}
{
  また、上腕を挙上させ、安定した後の様子を\figref{fig:estimate_average}に示す。
}
\switchlanguage{
  This shows that the position of the scapula is different when maximum muscle force is estimated from muscle volume and when it is averaged.
  When the maximum muscle force is averaged, the stable position of the scapula is shifted away from the spine.
  At \tabref{tab:maximum_muscle_forces}, the muscles with greater than average maximum muscle strength included the latissimus dorsi, pectoralis major, vastus lateralis, and subscapularis muscles.
  Among these muscles, the latissimus dorsi, pectoralis major, and subscapularis muscles all work to move the shoulder closer to the spinal column, so the smaller strength of these muscles is thought to have shifted the stable position of the shoulder blade away from the spine.
}
{
  これを見ると、最大筋力を筋体積から推定した時と平均化した時で、
  肩甲骨の位置が異なることがわかる。
  最大筋力を平均化した時は、肩甲骨の安定する位置が脊椎から遠ざかる方に移動している。
  \tabref{tab:maximum_muscle_forces}にて、最大筋力が平均よりも大きい筋肉として、
  広背筋、大胸筋、広頚筋、肩甲下筋等があった。
  このうち、広背筋、大胸筋、広頚筋はいずれも肩を脊柱に近づける向きに働く筋肉であるため、
  これらの筋力が小さくなったことで、
  肩甲骨の安定する位置が脊椎から遠ざかる方向に移動したと考えられる。
}

\switchlanguage{
  From these results,
  it was found that while the load on each muscle can be appropriately distributed by properly estimating the maximum muscle strength,
  conversely, if the maximum muscle strength is averaged, the muscles that are supposed to produce large force cannot exert sufficient force, and in order to raise the arm in that state, a large load is placed on other muscles, which also affects the posture in which the arm is raised.
}
{
  以上より、最大筋力を適切に推定することで各筋肉への負荷を適度に分散することができるが、
  逆に最大筋力を平均化すると、本来大きな力を出すはずの筋肉が十分な力を発揮できなくなり、
  その状態で腕を挙上させるために他の筋肉に大きな負荷がかかってしまい、
  挙上した姿勢にも影響を与えることがわかった。
}

\section{Conclusion}
\switchlanguage{
  In this study, ligaments were integrated into a musculoskeletal model of the human shoulder.
  The skeleton was treated as links, which were represented by mesh files scanned from a human skeleton, and structural joints were represented by spherical counterparts.
  Ligaments were implemented in spherical-dual joints primarily to limit the range of motion of the joints.
  The position of the ligaments was determined with reference to the arrangement of human ligaments, and these were represented by elastic wires.
  Muscles were placed based on human muscle mesh files, and their maximum muscle strength was estimated from muscle volume and length.  

  Although the model created was a multi-input system with 40 muscles in both shoulders, position control of the elbow tip was achieved by model predictive control that added signal input to the muscles to the evaluation function.
  Using positional control, we investigated the changes in posture by changing the presence or absence of ligaments and the distribution of muscle force in the upper arm raising movement.
  When the ligaments were removed, the joint range of motion was no longer restricted, the position of the scapula during raising was closer to the spine, and the angles of each bone during the raising motion showed unstable changes.
  In addition, when maximum muscle force was estimated from muscle volume and averaged, the load on each muscle was more distributed in the former case, and in the latter case, muscles with a larger ratio of muscle tension to maximum muscle force occurred.
  This may suggest that muscles that are thicker and have greater maximal muscle force are those that require greater force in human exercise.

  In the future, the developed simulation model could be used to examine how muscles should be used in more practical tasks in sports and martial arts by optimizing their movement.
  Also, in the design and control of musculoskeletal humanoids, it would be possible to reduce unnecessary output and price by using ligaments for joint configuration and by making its muscles high-power actuators for the larger muscles in the human body and low-power actuators for those that are not.
}
{
  本研究では、人間の肩を模した筋骨格モデルのシミュレータに靭帯を導入した。
  骨格をリンクとして扱い、これを人間の骨格をスキャンしたメッシュファイルによって表現し、
  構造的関節を球面対偶で表現した。
  靭帯は、球面対偶の関節において、主に関節の可動域を制限するために導入した。
  人間の靭帯の配置を参照しながら、靭帯の位置を決定し、これを弾性を持ったワイヤで表現した。
  筋肉は、人間の筋肉のメッシュファイルに基づいて配置し、その最大筋力を筋体積と長さから推定した。

  また、作成したモデルは両肩で40筋を有する多入力系であったが、
  筋肉への信号入力を評価関数に加えたモデル予測制御により肘先の位置制御を実現した。
  位置制御を利用し、上腕の挙上動作において靭帯の有無や筋力の分配を変えることによって、
  姿勢がどのように変化するかを検証した。
  靭帯をなくした場合、関節可動域の制約がなくなり、挙上時の肩甲骨の位置が脊椎に近づき、挙上動作中の各骨の角度が不安定な変化を示した。
  また、最大筋力を筋体積から推定した時と平均化した時で、前者のほうが各筋肉の負荷が分散されており、
  後者では最大筋力に対する筋肉の張力の比が大きくなる筋肉が生じた。
  このことは太く最大筋力の大きい筋肉は人間の運動において大きな力が要求される筋肉であることを示唆していると考えられる。

  今後は、開発したシミュレーションモデルを用いて、
  スポーツや武道のより実践的なタスクにおいて、
  その運動を最適化することでどのように筋肉が使われるべきかの考察が行えるだろう。
  また、筋骨格ヒューマノイドの設計や制御においても、
  関節の構成に靭帯を利用したり、その筋肉を人間の身体における大きい筋肉を高出力のアクチュエータにし、
  そうでない筋肉を低出力のアクチュエータにすることで、無駄な出力や価格を抑えることもできるだろう。
}

{
  \bibliographystyle{IEEEtran}
  \bibliography{bib}
}

\end{document}